\documentclass[conference,a4paper]{IEEEtran}
\IEEEoverridecommandlockouts

\usepackage[hidelinks]{hyperref}
\usepackage[cmex10]{amsmath}
\usepackage{amssymb,amsfonts}
\usepackage{dblfloatfix}

\usepackage[ruled,vlined]{algorithm2e}
\usepackage{graphicx}
\graphicspath{{Figures/PDF/}{Figures/PNG/}}

\usepackage{booktabs}
\usepackage{siunitx}
\usepackage[numbers,compress]{natbib}
\usepackage{texnames}
\usepackage{bm,bbm}
\usepackage{orcidlink}
\usepackage{multirow}

\begin{document}

\title{\uppercase{Cross-Polarization Fusion of VV and VH SAR Observations for Improved Flood Mapping}
\thanks{This work is supported by NASA award 80NSSC23M0051 and NSF Award 2401942. Copyright 2026 IEEE. Published in the 2026 IEEE International Geoscience 
and Remote Sensing Symposium (IGARSS 2026), scheduled for 9--14 August 
2026 in Washington, D.C. Personal use of this material is permitted. However, 
permission to reprint/republish this material for advertising or promotional 
purposes or for creating new collective works for resale or redistribution 
to servers or lists, or to reuse any copyrighted component of this work in 
other works, must be obtained from the IEEE. Contact: Manager, Copyrights 
and Permissions / IEEE Service Center / 445 Hoes Lane / P.O. Box 1331 / 
Piscataway, NJ 08855-1331, USA. Telephone: + Intl. 908-562-3966.

This version is the accepted manuscript submitted to arXiv. The final version 
will be published in the Proceedings of IGARSS 2026 and available via IEEE 
Xplore. For citation, please refer to the published version in IGARSS 2026.

\noindent * Correspondence to lhashemibeni@ncat.edu.}
}

\author{	\IEEEauthorblockN{Jagrati\ Talreja\orcidlink{0009-0009-4652-4196}}
	\IEEEauthorblockA{\textit{North Carolina A\&T State University}\\
		Greensboro, North Carolina, USA.\\
		jtalreja@ncat.edu}
\and
	\IEEEauthorblockN{Tewodros Syum Gebre\orcidlink{0000-0003-4508-2700}}
	\IEEEauthorblockA{\textit{North Carolina A\&T State University}\\
		Greensboro, North Carolina, USA.\\
		tsgebre@ncat.edu}
	\and
	\IEEEauthorblockN{Leila Hashemi Beni\orcidlink{0000-0003-1026-4555}}
	\IEEEauthorblockA{\textit{North Carolina A\&T State University}\\
		Greensboro, North Carolina, USA.\\
		lhashemibeni@ncat.edu}
}

\maketitle
\begin{abstract}
	Synthetic Aperture Radar (SAR) imagery is widely used for flood monitoring due to its all-weather and day–night imaging capability. However, flood mapping using single-polarization SAR data remains challenging in complex environments where surface and volume scattering coexist. In this paper, we investigate the effectiveness of cross-polarization fusion of VV and VH SAR observations for improved flood mapping. A deep learning–based segmentation framework is employed to jointly exploit complementary information from VV and VH polarizations. To ensure a fair evaluation, three configurations are compared under identical training conditions: VV-only, VH-only, and fused VV–VH input. Performance is assessed using standard flood mapping metrics, including Intersection over Union (IoU) and F1-score, along with qualitative visual analysis. Experimental results demonstrate that VV–VH fusion consistently outperforms single-polarization models, particularly in vegetated and heterogeneous flood regions, leading to more accurate flood boundary delineation. The findings highlight the importance of cross-polarization SAR fusion for enhancing the reliability of SAR-based flood mapping in disaster monitoring applications. \textit{https://github.com/JagratiTalreja01/Cross-Polarization-Fusion}
\end{abstract}


\section{Introduction}

Floods are among the most frequent and destructive natural disasters \cite{Hidalgo2019NaturalDisasters}, causing significant loss of life \cite{Jonkman2005GlobalFloodMortality}, infrastructure damage \cite{Diakakis2020FlashFloodTransport}, and economic disruption worldwide \cite{Blay2024FloodRiskIGARSS}. Timely and accurate flood mapping is essential for effective disaster response, risk assessment, and recovery planning \cite{Fawakherji2025FloodFusion}. Synthetic Aperture Radar (SAR) imagery has become a key data source for flood monitoring due to its ability to operate under all weather conditions and independent of daylight \cite{Chan2008IntroSAR}. Unlike optical sensors, SAR can reliably capture flood events during heavy rainfall and cloud cover \cite{Pierdicca2013FloodMappingSARWeather}, making it particularly valuable for rapid flood assessment \cite{Jamali2024ResidualWaveUNet}. However, the complex backscattering behavior of SAR signals poses challenges for accurate flood delineation \cite{Zhang2021ObjectBasedFloodSentinel1}, especially in heterogeneous landscapes containing vegetation, urban structures, and mixed land-cover types \cite{Pierdicca2017FloodModelsMethods}.

Traditional SAR-based flood mapping approaches often rely on thresholding techniques \cite{Liang2020LocalThresholdFloodSAR}, change detection \cite{Lu2015ObjectBasedChangeDetectionSAR}, or statistical modeling of backscatter intensity \cite{Martinis2015BackscatterFloodSAR} to distinguish flooded and non-flooded areas \cite{Zhao2024UrbanFloodSARRreview}. While effective in simple scenarios, these methods struggle in complex environments where surface roughness \cite{Anokye2024FloodResilienceWetlands}, vegetation structure \cite{Salem2022InundatedVegetationSAR}, and soil moisture produce similar backscatter responses to open water \cite{Pietroniro2005RemoteSensingWaterSoil}. To address these limitations, machine learning and deep learning techniques have been increasingly adopted for SAR-based flood mapping \cite{Andrew2023CNNFloodNovaSAR}, enabling the extraction of hierarchical features and improved discrimination of flood extents \cite{Fawakherji2025MultiResolutionFloodFusion}. Convolutional neural networks (CNNs), particularly U-Net–based architectures, have shown promising results by learning spatial and contextual information directly from SAR imagery \cite{Jamali2024ResidualWaveUNet}.

Despite these advances, many existing deep learning–based flood mapping methods rely on single-polarization SAR inputs, typically VV or VH, thereby underutilizing the complementary information provided by multi-polarization SAR observations \cite{Mohamadiazar2024NRTFloodDL}. VV polarization is generally more sensitive to surface scattering from open water and smooth surfaces, whereas VH polarization captures volume scattering effects from vegetation and partially submerged areas \cite{Deirmendjian1964ScatteringClouds}. Several studies have explored the use of multi-polarization SAR data for flood mapping; however, these approaches often employ simple feature concatenation or do not systematically evaluate the contribution of each polarization \cite{Qin2025HighPrecisionFloodSAR}. As a result, the potential benefits of explicitly fusing VV and VH information for robust flood segmentation remain insufficiently explored.

In this paper, we investigate the role of cross-polarization fusion of VV and VH SAR observations for improved flood mapping. We propose a deep learning–based segmentation framework that jointly exploits complementary scattering information from both polarizations. A controlled experimental design is adopted to compare VV-only, VH-only, and fused VV–VH configurations under identical network architectures and training settings, enabling a fair assessment of polarization fusion. Quantitative and qualitative evaluations demonstrate that VV–VH fusion leads to more accurate flood delineation, particularly in vegetated and heterogeneous regions. The results highlight the importance of cross-polarization SAR fusion for enhancing the reliability of SAR-based flood mapping in disaster monitoring applications.

\section{Methodology}

\subsection{Problem Formulation}
Flood mapping from Synthetic Aperture Radar (SAR) imagery is formulated as a supervised binary segmentation problem, where each pixel is classified as flooded or non-flooded. Dual-polarized SAR observations provide complementary scattering information: VV polarization is generally sensitive to surface scattering from open water and built-up structures, while VH polarization captures volume scattering effects associated with vegetation and partially inundated regions. Leveraging both polarizations can therefore improve flood delineation in heterogeneous landscapes. To further capture SAR-specific scattering behavior, ratio-based features (VV/VH and log(VV/VH)) are incorporated to enhance sensitivity to surface and volume scattering differences.

Given dual-polarized SAR inputs,
\begin{equation}
X = [X_{VV}, X_{VH}, X_{VV}/X_{VH}, \log(X_{VV}/X_{VH})] \in \mathbb{R}^{4 \times H \times W},
\end{equation}
the goal is to learn a mapping
\begin{equation}
F_\theta : \mathbb{R}^{2 \times H \times W} \rightarrow \mathbb{R}^{1 \times H \times W},
\end{equation}
where the output represents a pixel-wise flood probability map.

\begin{figure}[hbt]
	\centering
	\includegraphics[width=0.65\linewidth]{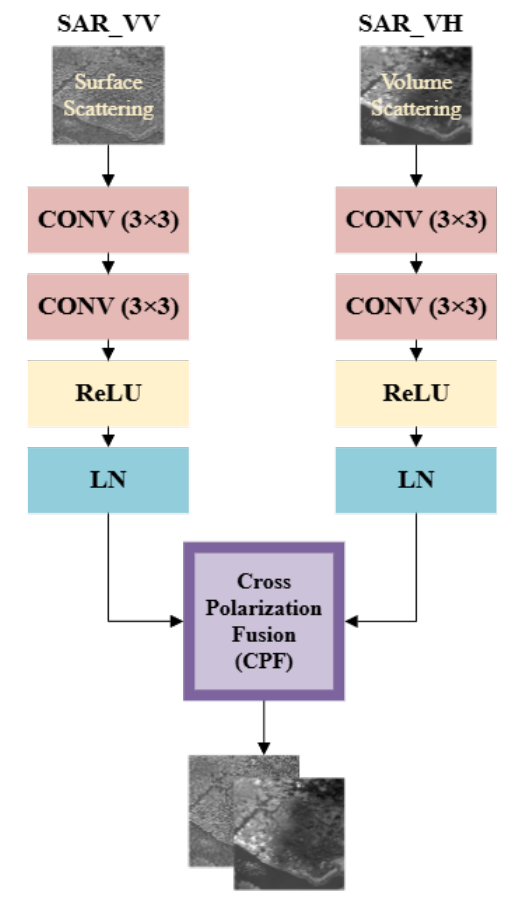}
	\caption{SAR\_VV \& SAR\_VH Polarization Fusion}\label{fig:Fusion1}
\end{figure}

\subsection{Polarization-Specific Feature Extraction}
To preserve distinct scattering characteristics, VV and VH channels are initially processed independently using lightweight convolutional stems as seen in Fig. 1:
\begin{equation}
F_{VV} = S_{VV}(X_{VV}), \quad F_{VH} = S_{VH}(X_{VH}),
\end{equation}
where $S_{VV}(\cdot)$ and $S_{VH}(\cdot)$ comprise stacked $3 \times 3$ convolutions followed by ReLU activation and layer normalization. This produces polarization-specific feature maps that are subsequently integrated prior to deeper encoding.

\subsection{Cross-Polarization Fusion (CPF)}
To effectively integrate complementary scattering cues, a Cross-Polarization Fusion (CPF) module is introduced. Unlike conventional early-fusion strategies such as channel concatenation, CPF adaptively emphasizes polarization-specific information using attention mechanisms guided by local spatial context. This design implicitly leverages the complementary scattering mechanisms of VV (surface scattering) and VH (volume scattering), enabling more effective flood discrimination in heterogeneous environments.

As shown in Fig. 2, the CPF module consists of two complementary components: Spatial Attention (SA) and Channel Attention (CA). For each polarization feature map, spatial attention highlights informative regions by learning spatial importance weights, while channel attention selectively emphasizes feature channels relevant to flood discrimination. Spatial attention is computed using pooled spatial descriptors followed by convolution and sigmoid activation, whereas channel attention is derived from global pooling operations and a shared multilayer perceptron.

Cross-polarization interaction is achieved by enabling bidirectional attention between VV and VH features. VV features attend to VH features to enhance flood-related cues in vegetated and mixed land-water regions, while VH features attend to VV features to reinforce structural boundaries and open-water regions. The attention-weighted features from both directions are concatenated and projected using a $3 \times 3$ convolution:
\begin{equation}
F_{\text{fused}} = \text{Conv}_{3 \times 3} \left( \left[ F_{VV}^{att} \parallel F_{VH}^{att} \right] \right),
\end{equation}
where $[\cdot \parallel \cdot]$ denotes channel-wise concatenation.

\begin{figure}[hbt]
	\centering
	\includegraphics[width=\linewidth]{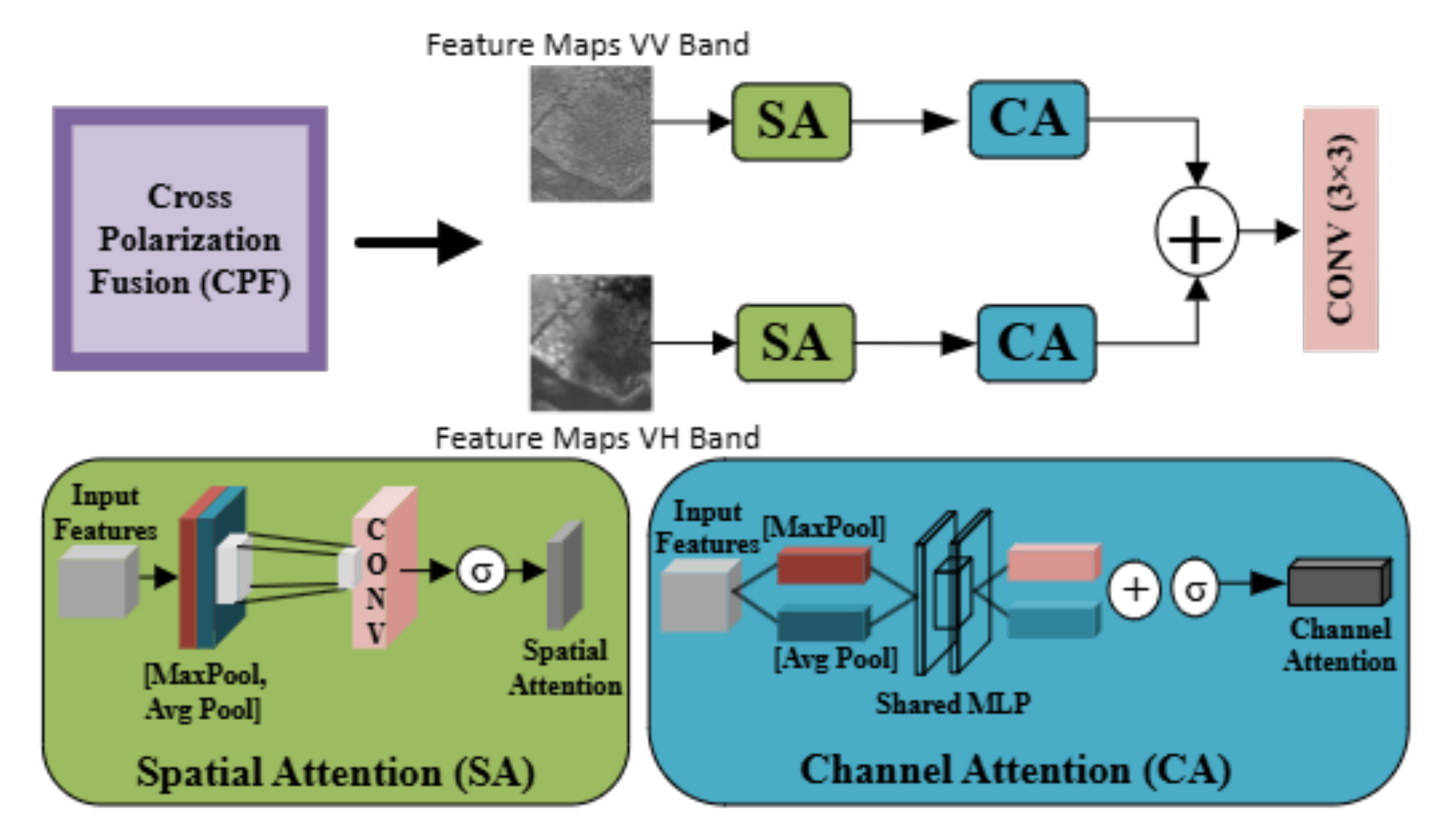}
	\caption{Cross Polarization Fusion (CPF)}\label{fig:Fusion2}
\end{figure}

\subsection{Networks}
We employ standard encoder-decoder segmentation models as baselines, specifically (i) a U-Net with skip connections and (ii) a convolutional autoencoder without skip connections. These architectures are selected to ensure a controlled and fair evaluation of the proposed fusion strategy, independent of backbone complexity. Both networks follow the common structure of progressively downsampling feature maps to capture context and upsampling to recover pixel-level predictions. The final layer uses a sigmoid activation to output a probabilistic flood map, which is thresholded to obtain a binary mask. \textit{CPF is inserted at the input stage and kept identical for both U-Net and autoencoder to isolate the effect of cross-polarization fusion}.

\subsection{Flood Segmentation and Training Strategy}
The fused feature representation is passed through the encoder--decoder network to produce pixel-level flood predictions. A sigmoid activation function is applied to generate a probabilistic flood map, which is thresholded to obtain the final binary segmentation. The network is trained in a supervised manner using binary cross-entropy loss. To ensure a fair evaluation of polarization fusion, all comparative experiments (VV-only, VH-only, and VV--VH fusion) employ identical architectures, training parameters, and data splits. Data augmentation techniques such as random flipping and rotation are applied consistently across all experiments to improve generalization. Performance is assessed using Intersection over Union (IoU) and F1-score, along with qualitative visual inspection of flood boundary delineation.

\section{Experiments \& Results}

\subsection{Experimental Setup}

Experiments are conducted using dual-polarized SAR imagery acquired in VV and VH polarizations, along with corresponding reference flood masks. All SAR images are preprocessed to ensure spatial alignment between polarizations and ground truth labels. The data are divided into training, validation, and test sets with no spatial overlap to prevent information leakage. Image patches of fixed size are extracted to facilitate batch training.

Two baseline segmentation networks are evaluated: a U-Net with skip connections and a convolutional autoencoder without skip connections. For each network, three input configurations are considered: (i) VV-only, (ii) VH-only, and (iii) fused VV–VH input using the proposed Cross-Polarization Fusion (CPF) module. Except for the input configuration, all networks share identical architectures, training parameters, and data splits to ensure a fair comparison.

All models are trained using binary cross-entropy loss and optimized using the same optimizer and learning rate schedule. Standard data augmentation techniques, including random horizontal and vertical flipping, are applied consistently across all experiments.

\subsection{Evaluation Metrics}

Flood mapping performance is evaluated using standard pixel-wise segmentation metrics, including Intersection over Union (IoU), F1-score, and Overall Accuracy (OA). These metrics are computed based on the number of true positives (TP), false positives (FP), false negatives (FN), and true negatives (TN) obtained from the predicted and reference flood masks. IoU is equivalent to the Critical Success Index (CSI), a widely used metric in flood mapping.

Intersection over Union (IoU), also known as the Jaccard Index, measures the overlap between the predicted flood region and the ground truth:
\begin{equation}
\text{IoU} = \frac{TP}{TP + FP + FN}.
\end{equation}

The F1-score evaluates the balance between precision and recall and is defined as:
\begin{equation}
\text{F1-score} = \frac{2TP}{2TP + FP + FN}.
\end{equation}

Overall Accuracy (OA) represents the proportion of correctly classified pixels:
\begin{equation}
\text{OA} = \frac{TP + TN}{TP + TN + FP + FN}.
\end{equation}

These metrics provide complementary perspectives on flood segmentation performance, with IoU and F1-score emphasizing flood extent delineation and robustness to class imbalance, while OA reflects overall pixel-level classification accuracy.

\subsection{Dataset}

We use the DeepFlood dataset \cite{Fawakherji2025DeepFloodDataset}, which contains dual-polarized Sentinel-1 SAR imagery (VV, VH) and corresponding flood masks from two major events in North Carolina: Hurricane Matthew (2016) and Hurricane Florence (2018). The data are radiometrically calibrated and co-registered, and divided into fixed-size patches for training. The Sentinel-1 data are dual-polarized (VV, VH) with 10 m spatial resolution, which may limit fine-scale flood detection.

A cross-event evaluation strategy is adopted to assess generalization. Data from Hurricane Florence are split into 80\% for training and 20\% for testing, while Hurricane Matthew is used entirely for testing as an unseen event. This setup evaluates model robustness across different flood conditions and acquisition times. Table~\ref{tab:dataset} summarizes the dataset and split details.
Table~\ref{tab:dataset} summarizes the dataset composition and data split strategy used in this study.

\begin{table}[t]
\centering
\caption{Dataset details and data split strategy.}
\label{tab:dataset}
\resizebox{0.5\textwidth}{!}{
\begin{tabular}{lcccccc}
\hline
\textbf{Hurricane} & \textbf{Sensor} & \textbf{Polarization} & \textbf{Resolution} & \textbf{Region} & \textbf{Train} & \textbf{Test} \\
\hline
Matthew & Sentinel-1 & VV, VH & 10 m & Grifton & -- & 100\% \\
Florence & Sentinel-1 & VV, VH & 10 m & Bladen & 80\% & 20\% \\
\hline
\end{tabular}
}
\end{table}

\subsection{Results}

Quantitative and qualitative results are presented to evaluate the impact of cross-polarization fusion on SAR-based flood mapping. Performance is compared across three input configurations: VV-only, VH-only, and fused VV–VH using the Cross-Polarization Fusion (CPF) module. All results are reported on the test sets using identical training and evaluation protocols to ensure a fair comparison.

\subsubsection{Quantitative Results}

Table II and Table III summarizes the flood segmentation performance in terms of Intersection over Union (IoU) and F1-score. Across both networks, U-Net and the convolutional autoencoder, the fused VV–VH configuration consistently outperforms single-polarization inputs. Models trained with VV-only inputs generally perform better than VH-only models in open water regions due to stronger surface scattering contrast. However, VH-only models exhibit improved sensitivity in vegetated and mixed land–water areas, highlighting the complementary nature of the two polarizations.

The introduction of CPF leads to notable improvements in both IoU and F1-score, indicating more accurate flood extent delineation. These gains are achieved without modifying the depth or complexity of the baseline networks, confirming that performance improvements are primarily attributed to effective cross-polarization fusion rather than architectural changes.

\begin{table}[t]
\centering
\caption{Flood segmentation performance using U-Net (IoU/CSI \% / F1-score \%).}
\label{tab:unet_results}
\begin{tabular}{lccc}
\hline
\textbf{Method} & \textbf{IoU} & \textbf{CSI} & \textbf{F1-score} \\
\hline
VV only & 66.2 & 66.2 & 79.7 \\
VH only & 62.5 & 62.5 & 76.9 \\
Addition Fusion & 67.4 & 67.4 & 79.9 \\
Early Fusion (Concat) & 68.1 & 68.1 & 80.5 \\
CPF (VV, VH) & \textbf{69.8} & \textbf{69.8} & \textbf{82.2} \\
\hline
\end{tabular}
\end{table}

\begin{table}[t]
\centering
\caption{Flood segmentation performance using autoencoder (IoU/CSI \% / F1-score \%).}
\label{tab:ae_results}
\begin{tabular}{lccc}
\hline
\textbf{Method} & \textbf{IoU} & \textbf{CSI} & \textbf{F1-score} \\
\hline
VV only & 60.4 & 60.4 & 75.3 \\
VH only & 57.1 & 57.1 & 72.7 \\
Addition Fusion & 61.1 & 61.1 & 75.8 \\
Early Fusion (Concat) & 61.8 & 61.8 & 76.4 \\
CPF (VV, VH) & \textbf{63.2} & \textbf{63.2} & \textbf{77.5} \\
\hline
\end{tabular}
\end{table}

\subsubsection{Qualitative Analysis}

Fig. 3 and Fig. 4 presents representative qualitative flood mapping results for different polarization inputs. Single-polarization models often produce fragmented flood masks or misclassify flooded vegetation, particularly along riverbanks and floodplain boundaries. In contrast, the CPF-based VV–VH fusion yields more coherent flood regions and sharper boundary delineation.

\begin{figure}[hbt]
	\centering
	\includegraphics[width=\linewidth]{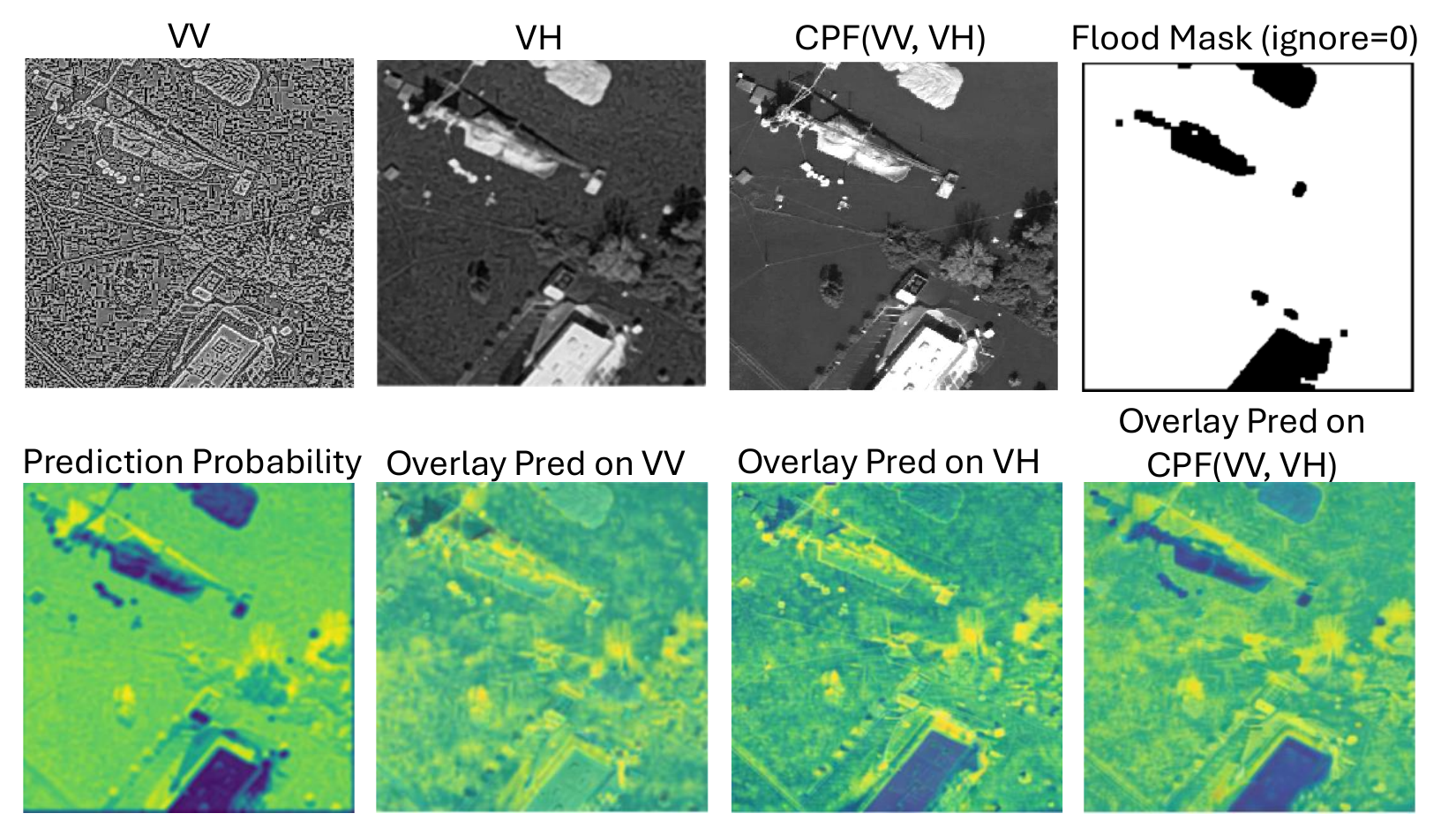}
	\caption{Comparison of flood mapping results obtained using UNet with VV-only, VH-only, and CPF (VV,VH).}\label{fig:ResultUNet}
\end{figure}

\begin{figure}[hbt]
	\centering
	\includegraphics[width=\linewidth]{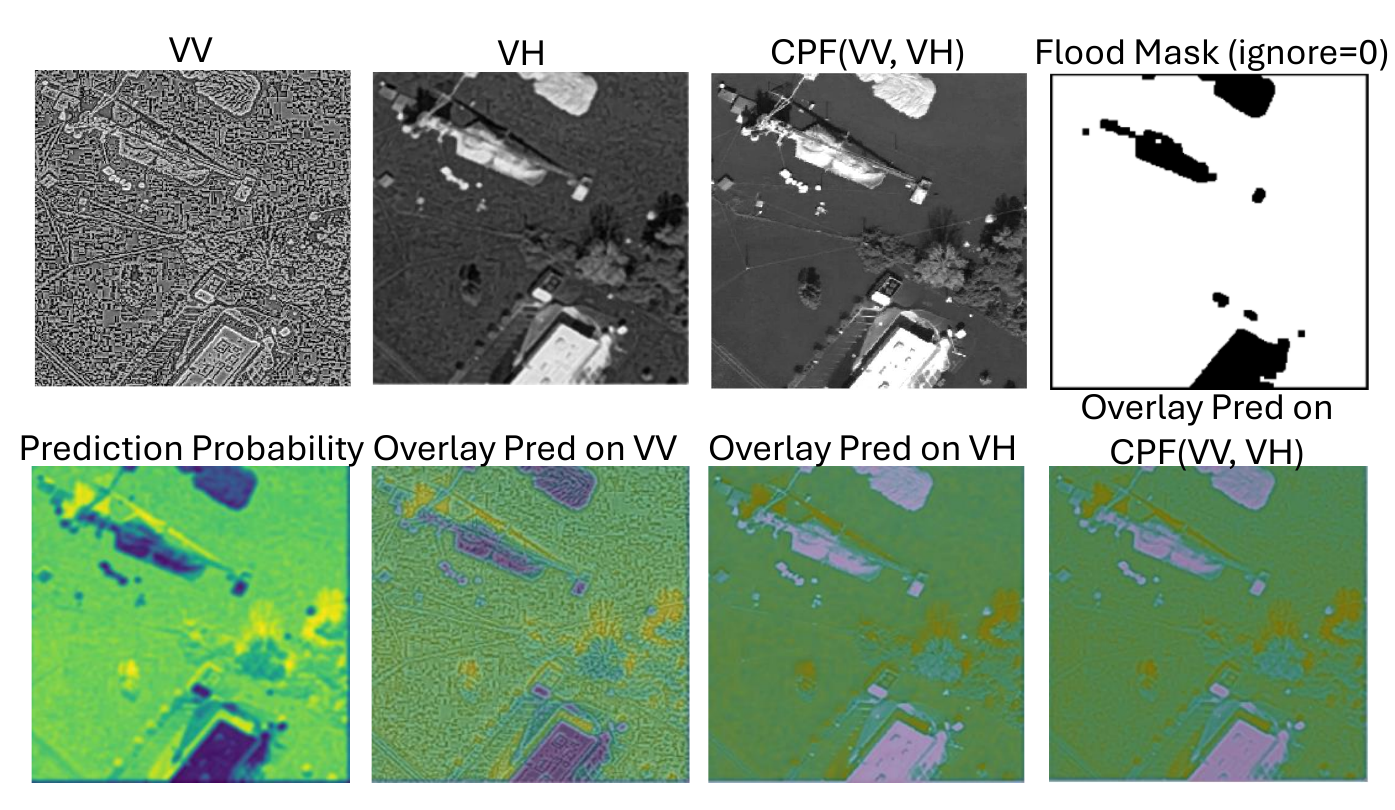}
	\caption{Comparison of flood mapping results obtained using Auto-Encoder with VV-only, VH-only, and CPF (VV,VH).}\label{fig:ResultAE}
\end{figure}

The fused results demonstrate improved robustness in heterogeneous environments, where surface and volume scattering effects coexist. By adaptively integrating VV and VH information, the CPF module enhances flood detection in challenging regions that are poorly represented by either polarization alone.
Evaluation on the Hurricane Matthew dataset, which is excluded from training, further demonstrates the generalization capability of the proposed fusion strategy. While single-polarization models show noticeable performance degradation under unseen flood conditions, the VV–VH fusion model maintains more stable segmentation performance, indicating improved robustness to variations in flood dynamics and acquisition conditions.

\section{Conclusion}
This study evaluated dual-polarization Sentinel-1 SAR data for flood mapping, comparing single-polarization, early-fusion, and attention-based fusion strategies. Results demonstrate that VV–VH fusion improves flood segmentation performance, highlighting the importance of explicitly modeling complementary scattering mechanisms. The proposed CPF module achieves consistent gains, indicating its effectiveness for robust flood mapping in heterogeneous environments. 
Future work will explore integration with advanced architectures (e.g., SegFormer, Swin-UNet) and incorporate uncertainty analysis to further enhance model reliability.

\section*{Acknowledgments}
This work is supported by NASA award 80NSSC23M0051 and NSF Award 2401942.

\bibliographystyle{IEEEtranN}
\bibliography{references}

\end{document}